# Les Entités Nommées : usage et degrés de précision et de désambiguïsation

Claude Martineau, Elsa Tolone, Stavroula Voyatzi[1]
**Université Paris-Est Marne-la-Vallée**

**Abstract**

The recognition and classification of Named Entities (NER) are regarded as an important component for many Natural Language Processing (NLP) applications. The classification is usually made by taking into account the immediate context in which the NE appears. In some cases, this immediate context does not allow getting the right classification. We show in this paper that the use of an extended syntactic context and large-scale resources could be very useful in the NER task.

## Introduction

La nécessité de maîtriser des flux et des stocks d'informations multimédias en croissance constante affecte les secteurs les plus variés de l'économie et de l'action gouvernementale. L'information d'aujourd'hui est en effet :

- **massive** : le Web visible représente aujourd'hui 8 milliards de pages, les disques durs des particuliers et des entreprises environ le centuple. On parle aujourd'hui de Gigaoctets voire de Téraoctets d'informations (ou de données) disponibles ;

- **complexe et hétérogène** : il ne s'agit plus seulement aujourd'hui de gérer des données qui se présentaient sous forme structurée dans des fichiers plats ou dans de petites bases de données, mais bien de traiter et d'analyser des fichiers plus complexes comme des textes non structurés, des images et même des enregistrements audio et vidéo ;

- **soumise à des contraintes de temps réel** : la prise de décision s'effectue aujourd'hui habituellement sous le signe de l'urgence ; la « crise » constituant le cadre quasi normal de l'action aussi bien dans l'entreprise que dans le domaine gouvernemental.

Créer des outils qui automatisent l'exploration ou l'extraction d'informations pertinentes, notamment dans les textes, est alors crucial. Les systèmes d'extraction d'information (Hobbs *et al.* 1996 ; Fourour 2002), de recherche d'information (Sekine et Isahara 1998) et de fouille de textes (Jacquemin et Bush 2000) sont de plus en plus nombreux. La tâche d'extraction d'information (en anglais *Information Extraction*, désormais IE) consiste à repérer dans les documents les informations permettant de convertir le texte en une fiche signalétique standard à partir de laquelle il est possible de répondre à des questions factuelles telles que : *quand* et

---

[1] Institut Gaspard-Monge, Université Paris-Est Marne-La-Vallée, {claude.martineau ; elsa.tolone ; stavroula.voyatzi}@univ-mlv.fr.

*où se déroule(nt) le ou les évènements relatés dans le texte ? quels sont les acteurs et quel est leur rôle ?...* Il est possible de voir cette tâche comme conversion d'informations faiblement structurées en données structurées, qui seront alors plus faciles à exploiter (Li *et al.* 2002).

Notre travail s'inscrit dans le cadre d'un projet de recherche et s'intéresse à l'extraction des entités nommées (désormais EN) dans un corpus bien spécifique (section 2). Après une présentation de l'objectif de la tâche d'extraction d'EN (section 1) précisant le rôle des ressources logicielles et linguistiques utilisées, nous nous penchons sur les problèmes liés à la désambiguïsation et au raffinement de la catégorisation d'EN (section 3). Enfin, nous présentons nos résultats (section 4) et les perspectives qu'ouvre ce travail.

## 1. Entités nommées

Selon (Chinchor 1998), la notion d'EN est utilisée pour regrouper tous les éléments du langage qui font référence à une entité unique et concrète, appartenant à un domaine spécifique (*ie.* humain, économique, géographique, etc.). On désigne traditionnellement par le terme d'EN les noms propres au sens classique, les noms propres dans un sens élargi mais aussi les expressions de temps et de quantité. Le terme d'« entité nommée » est utilisé dans le domaine du traitement automatique des langues (TALN) depuis les conférences *Message Understanding Conferences* (MUC[2]) qui ont établi la catégorisation suivante :

| Catégories | Sous-catégories |
|---|---|
| Named Entities (ENAMEX) | « Person », « Organization », « Location » |
| Temporal Expressions (TIMEX) | « Date », « Time » |
| Number Expressions (NUMEX) | « Money », « Percent » |

La reconnaissance et l'extraction d'EN sont des tâches bien établies depuis les conférences MUC. Repérer et catégoriser les EN permet un accès particulièrement pertinent au contenu des documents et, de ce fait, représente un enjeu crucial pour l'analyse et la compréhension automatique des textes. De nombreux systèmes ont été développés pour l'extraction d'EN en différentes langues. Certains reposent sur une approche symbolique et utilisent une base de règles construites à la main (Hobbs *et al.* 1996). D'autres sont fondés sur une approche statistique et emploient des connaissances acquises automatiquement par apprentissage (Bikel *et al.* 1997). D'autres, enfin, adoptent une approche hybride (Watrin 2006).

Cependant, les premières définitions et catégorisations des EN, selon les conférences MUC, ne permettent pas de répondre aux besoins des différentes applications : certaines se contenteront d'annotations classiques telles que "lieu", "organisation", "personne" tandis que d'autres auront besoin d'annotations beaucoup plus fines. Un outil de veille technologique aura par exemple besoin de distinguer les occurrences de *Renault* en tant que « société » des occurrences de *Renault* en tant que « véhicule ». En revanche, le même outil n'aura pas besoin de savoir que *Arnold Schwarzenegger* peut avoir différentes fonctions telles que « acteur », « gouverneur de Californie » ou « bodybuilder ». Ce sera l'inverse pour un outil de veille politique. D'autre part, les difficultés d'annotation liées à l'ambiguïté des entités se retrouvent quelque soit le degré de finesse de catégorisation recherché. Du point de vue linguistique, les problèmes rencontrés sont de type homonymie-polysémie (Kleiber 1999) ou de type métonymique (e.g. **La France** *a signé le traité de Kyoto* : « pays » ou « gouvernement » ?). Certains cas sont relativement aisés à résoudre, mais d'autres posent davantage problème, même pour un humain (e.g. **Orange** *a invité M. Dupont* : « ville » ou

---

[2] Voir le site : http ://www.muc.saic.org.

« société » ?). Ainsi, si l'on veut caractériser plus finement les EN, il importe avant tout de les désambiguïser. Après avoir considéré plus avant les divers problèmes liés au statut référentiel des EN (Poibeau 2005), nous présentons la catégorisation établie pour les besoins du projet.

## 2. Etude expérimentale

Notre étude s'inscrit dans le cadre du projet de Recherche et Développement Infom@gic[3] qui concerne le domaine de l'Analyse de l'Information. Le projet vise à mettre en place un laboratoire industriel de sélection, de tests, d'intégration et de validation d'applications opérationnelles des meilleures technologies franciliennes dans le domaine de l'ingénierie des connaissances. Le présent travail porte sur l'extraction et l'annotation fine de certains types d'EN dans les documents textuels.

### 2.1. Corpus d'étude

Notre recherche a été réalisée sur un corpus interne de THALES L&J, coordinateur du projet Infom@gic. Il s'agit d'un corpus brut en français de 38 Mégaoctets comportant 3 173 071 occurrences de mots. Il est constitué de dépêches d'agences et d'extraits de presse portant sur les événements politiques en Côte d'Ivoire et au Kossovo durant la période 2000-2003.

### 2.2. Typologie des EN

Plusieurs typologies ont été proposées pour les noms propres en général ; certaines morpho-syntaxiques (Allerton 1987), d'autres graphiques (Daille et Morin 2000) ou bien encore sémantiques (Grass 2000). Pour ce qui est des EN, d'autres catégorisations ont été établies, parmi lesquelles nous citons ESTER (Le Meur, Galliano et Geoffrois 2004) et celle de (Sekine *et al.* 2002). La typologie adoptée dans le projet comporte les neuf classes d'EN suivantes :

| | | |
|---|---|---|
| Noms de personnes | Noms de lieux (expressions spatiales incluses) | Noms de faits |
| Noms d'organisations | Dates et heures (expressions temporelles incluses) | Noms de moyens |
| Coordonnées | Expressions numériques | Noms d'œuvres |

Chacune de ces classes est affinée par un ensemble de sous-classes et est illustrée par le biais d'un schéma, qui spécifie : la classe de l'EN, ses types et sous-types ainsi que ses attributs. La Figure 1. illustre le schéma de représentation des EN de type « Noms de lieux ». Elle est suivie par l'ensemble des types, sous-types et attributs associés à cette classe d'EN.

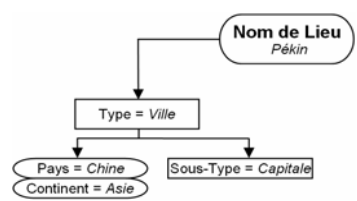

*Figure 1. EN de type « Lieu »*

**Types et sous-types**
- Groupe de pays
    - Regroupement géographique (ie. *Pays de l'Afrique de l'Ouest*)
    - Regroupement économique (ie. *Union économique et monétaire ouest-africaine*, *UEMOA*)
- Continent (*ie. Afrique*)
- Pays (*ie. Côte d'Ivoire*)
- Etat (*ie. Washington*, État fédéral des Etats-Unis)
- Région (*ie. Ile-de-France*)
- Capitale (*ie. Vienne*, capitale de l'Autriche)

---

[3] Pour plus d'informations, voir : https://www4.online.thalesgroup.com/QuickPlace/infomagic/Main.nsf.

- Département (*ie. Vienne*, département de France)
- Ville (*ie. Lyon*, ville de France)
- Microtoponyme
    Place, Aéroport, …
- Hydronyme
    Lac (*ie. lac Baïcal*), Fleuve (*ie. La Seine*), …
- Oronyme
    Montagne (*ie. Le Mont Blanc*, France)
    Désert (*ie. désert du Niger*, Niger), …
- Voies de communication
    Rue (ie. *rue de Montreuil*, Paris), Boulevard (ie. *boulevard Haussmann*, Paris), …

**Attributs :**

- Pays
- Ville
- Numéro de département
- Langue officielle
- Capitale
- Superficie
- Nombre d'habitants

Certaines classes, types et sous-types d'EN ne sont pas utiles dans toutes les applications, alors que de nouveaux peuvent être nécessaires suivant l'apparition de différents besoins. Cette typologie doit donc être ouverte et paramétrable pour chaque application. Nous allons à présent décrire la méthode et les ressources utilisées dans notre travail.

## 3. Méthode et ressources utilisées

Le but de notre travail est de développer un système d'extraction et d'annotation fine d'EN en réutilisant autant que possible les ressources et outils existants, en particulier, ceux développés au sein de l'équipe d'Informatique Linguistique de l'IGM[4], à l'Université de Marne-la-Vallée. Nous avons donc adopté une approche symbolique qui nous a également permis de valider la couverture et la pertinence de nos ressources, ainsi que de les adapter pour passer de la simple reconnaissance à l'extraction. Cette méthode a déjà été expérimentée dans le domaine d'IE (Roche et Schabes 1997 : 149-168, 329-352, 383-402) et s'est révélée suffisamment efficace. Pour l'extraction des EN, nous avons fait appel au système Unitex (Paumier 2003). Unitex est un environnement de développement qui permet de construire des descriptions formalisées de grammaires et d'utiliser des ressources telles que des dictionnaires généraux et spécialisés. Tous les objets traités par Unitex sont ou peuvent être transformés en des transducteurs à nombre fini d'états (*RTN* en anglais).

### 3.1. L'utilisation des dictionnaires électroniques

La constitution de dictionnaires électroniques est largement utilisée pour l'extraction des EN. Ces dictionnaires peuvent être spécialisés et comportent alors des EN en tant que telles (*ie.* anthroponymes, toponymes, sigles), ou généraux et contiennent des mots figurant en tant que contexte immédiat d'une EN, qu'ils en fassent partie ou non (*ie.* la **ville** de Marseille). Les dictionnaires spécialisés que nous avons utilisés dans notre tâche sont les suivants :

| Type de dictionnaire | Auteur | Exemple |
|---|---|---|
| Prénoms | Maurel *et al.* 1996[5] | *Caroline,.N+PR+Hum+Prénom:fs* |
| Toponymes | Maurel et Piton 1999 | *Seine,.N+PR+Hydronyme:fs* |
| Pays, Capitales et Gentilés | Maurel et Piton 1999 | *France,.N+PR+Toponyme+Pays+IsoFR:fs* |
| Adjectifs toponymiques | Maurel et Piton 1999 | *parisiens,parisien.A+Toponyme+Ville:mp* |
| Noms de profession | Fairon 2004 | *banquiers,banquier.N+Profession:mp* |
| Sigles et Abréviations | Maurel *et al.* 1996 | *Solensi,Solidarité Enfants Sida.N+Sigle:fs* |

---

[4] Voir : http://infolingu.univ-mlv.fr/.

[5] Les dictionnaires ont été élaborés dans le cadre du projet Prolex.

Nous avons également utilisé les dictionnaires morphologiques des mots simples (984 723 entrées) et composés (272 228 entrées) du français (Courtois 1990). Enfin, nous avons enrichi semi-automatiquement, à partir du corpus et du Web, les ressources déjà existantes d'environ 500 entrées (ie. « Noms d'Organisations » et « Abréviations », Tolone 2006 : 16).

3.2. L'exploitation du contexte immédiat : l'utilisation de grammaires locales

La création de grammaires locales sous forme de graphes *RTN* est amplement utilisée pour l'extraction des EN. Cette méthode consiste à utiliser la présence de mots appelés « mots déclencheurs » dans le contexte immédiat (droit ou gauche) d'une EN potentielle.

On peut donc reconnaître l'EN *Neuf **Télécom*** grâce à la présence du déclencheur *Télécom* qui constitue, selon (McDonald 1996 : 22), une preuve interne car il fait partie de l'EN elle-même. De manière similaire, on reconnaît que *Vivendi*, dans le contexte *Le **groupe** Vivendi prend le contrôle de Neuf Télécom*, est une EN de type « Organisation ». Selon (McDonald 1996 : 23), le déclencheur *groupe* représente ici une preuve externe car il n'appartient pas forcément à l'EN et constitue le contexte d'apparition de l'EN dans l'occurrence donnée.

Les grammaires locales que nous avons utilisées comportent environ 600 graphes. Ces graphes ont été élaborés au sein de l'IGM depuis de nombreuses années et sont rassemblés et accessibles grâce au système Graalweb (Constant 2004). Initialement prévus pour effectuer la reconnaissance de patterns morphosyntaxiques, ils ont dû être adaptés pour l'extraction d'EN. Ceci a consisté à transformer ces automates en transducteurs qui produisent les balises initiales et finales délimitant chaque EN. De plus, des variables y ont été ajoutés afin d'extraire non seulement l'EN mais aussi les attributs la concernant (cf. Figure 2.). Ceci permet de fournir comme résultat de l'extraction une fiche où figure l'occurrence de l'EN reconnue assortie de l'ensemble de ses attributs présents dans l'occurrence (cf. Figure 3.).

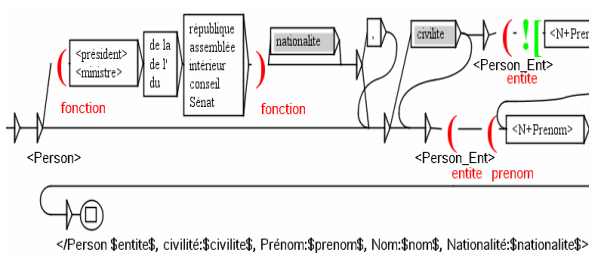 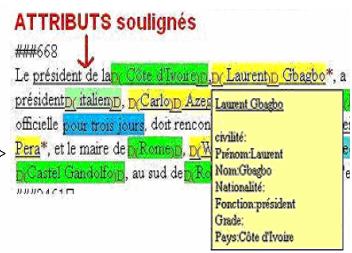

*Figure 2.* *Figure 3.*

*Graphe reconnaissant des EN de type « Personne »* *Visualisation sous forme de fiches*

La preuve externe est nécessaire pour obtenir des performances élevées dans l'extraction des EN (Friburger 2002). Si on ne prend en compte que la preuve interne, on peut souvent aboutir à des erreurs de catégorisation. Par exemple, le nom propre contenu dans *la société Hugues Aircraft* pose problème : *Hugues* figure dans le dictionnaire des prénoms. La seule preuve interne apportée par ce prénom fait penser que *Hugues Aircraft* est un « nom de personne », ce qui est contredit par la preuve externe *société*. Ce type d'erreur de catégorisation est très fréquent entre « noms de personnes » et « noms d'organisations ».

3.3. L'exploitation du contexte éloigné : le cas du contexte syntaxique droit

Les contextes pris généralement en compte ne font intervenir que des noms ou adjectifs. La prise en compte de contextes plus longs et d'autres catégories grammaticales, notamment les verbes et les expressions figées, est essentielle pour ne pas aboutir à une catégorisation erronée de l'entité traitée. Considérons l'exemple :

*Le (((Quai d'(Orsay)) se trouve) dans l'impossibilité) d'affirmer que…*

[Dictionnaire]  (Orsay,.N+PR+Toponyme+Ville:fs)
[Preuve interne]  (Quai d'Orsay : lieu_micro-toponyme)
[Preuves interne/externe]  (Quai d'Orsay_se trouve : lieu_micro-toponyme)
[Contexte éloigné]  (Quai d'Orsay_se trouve dans l'impossibilité : organisation)

La première erreur de reconnaissance (Orsay : lieu_ville) est fréquente dans la mesure où les EN de lieu sont souvent imbriquées dans des constituants plus grands qui sont eux-mêmes des EN. La deuxième erreur de reconnaissance est due à la prise en compte de la preuve interne (présence du déclencheur *Quai*), qui nous amènerait à fournir la catégorie « lieu_micro-toponyme ». La prise en compte du contexte syntaxique droit *se trouve* ne nous permet pas non plus de corriger l'étiquetage car ce verbe peut accepter un nom de lieu en position de sujet. Seule la prise en compte du complément prépositionnel *dans l'impossibilité* nous amènerait à la catégorisation correcte de l'EN. *Quai d'Orsay* peut donc être correctement reconnu grâce à l'exploitation du contexte syntaxique éloigné, en l'occurrence le prédicat complexe *N0 Vsup Prép X=: N0 se trouver dans l'impossibilité de V0inf W* VS *N0 se trouver Loc N*, sans qu'il soit besoin de faire appel à la notion de référent (Quai d'Orsay= Ministère des Affaires Etrangères) ni à des ressources extérieures (*ie.* bases de connaissances, encyclopédies, etc.).

## 4. Résultats et problèmes

Nous avons reconnu dans ce corpus 189 021 occurrences d'EN réparties dans les catégories suivantes : 49 135 EN de type personne, 21 765 EN de type organisation, 60 766 EN de type lieu, 37 520 EN de type date, 9 685 EN de type durée et 10 150 EN de type heure.

Les résultats obtenus par application des grammaires locales sont traités par un programme qui effectue un traitement en deux phases nécessitant chacune une lecture du texte. Lors de la première phase, les EN reconnues sont mémorisées dans un lexique. On recherche également dans le texte les séquences de mots qui commencent par une majuscule et qui ne sont pas reconnues. Ces séquences sont étiquetées « NON RECONNU ». Lors de la seconde phase, ces séquences sont recherchées parmi les EN déjà mémorisées. Après comparaison, celles qui sont des occurrences d'une EN déjà mémorisée sont étiquetées de la même manière. Ceci permet d'accroître le taux de reconnaissance des EN. Les changements de fréquences dans le lexique en témoignent (cf. Figure 4).

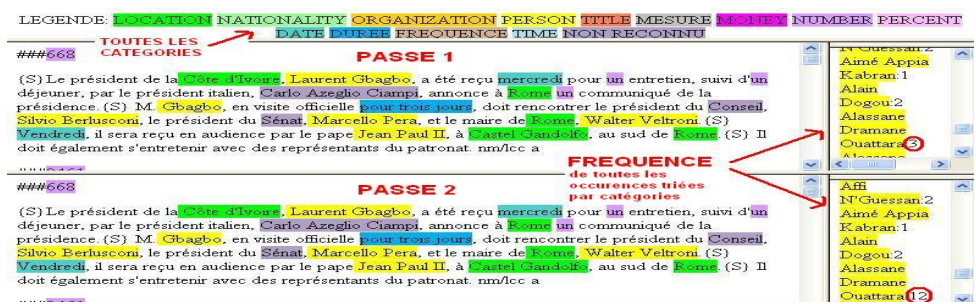

*Figure 4. Visualisation des EN reconnues par double lecture*

Cette étude étant en cours et faute d'un corpus de référence (annoté à la main), nous ne pouvons pas parler encore de précision et de rappel. En revanche, il nous est possible d'attribuer des étiquettes « plus fines » et correctes pour certaines catégories d'EN (cf. §3.3). D'ailleurs, ce corpus possède un certain nombre de particularités qui ont rendu cette tâche plus complexe : taille importante des dépêches (allant jusqu'à plusieurs milliers de caractères), (micro-)toponymes africains absents des dictionnaires, noms propres africains comportant des apostrophes infixes (*i.e. N'Djamena*), séquences isolées couramment utilisées (*i.e. Éternel **Gbagbo** !*) et titres des articles entièrement en majuscules.

## Conclusion et perspectives

Cette étude a montré la validité des grammaires locales existantes au sein du système Graalweb. Leur transformation pour passer de la stricte reconnaissance des EN à leur extraction a été une partie substantielle de ce travail. Cette étude a également mis en relief l'importance qu'il faut accorder à des ressources linguistiques adaptées sans lesquelles une catégorisation correcte des EN ne saurait être obtenue. D'ailleurs, les catégories grammaticales (nom ou adjectif) habituellement prises en compte comme déclencheurs se révèlent souvent insuffisantes et aboutissent à une catégorisation erronée. L'extension du contexte d'analyse et la prise en compte du verbe et des phénomènes de figement, notamment en ce qui concerne le contexte syntaxique droit, sont fort souhaitables. Dans ce but, nous envisageons dans la suite de ce travail d'utiliser les ressources présentes au sein du lexique-grammaire (Gross 1981) afin d'améliorer la précision de l'extraction des EN.

## Références